\documentclass[10pt,twocolumn,letterpaper]{article}

\usepackage{wacv}
\usepackage{times}
\usepackage{epsfig}
\usepackage{graphicx}
\usepackage{amsmath}
\usepackage{amssymb}
\usepackage{subcaption}    
\usepackage{booktabs}       
\usepackage{authblk}          
\usepackage[bookmarks=false]{hyperref}         

\wacvfinalcopy 

\def\wacvPaperID{962} 

\ifwacvfinal\fi
\begin{document}

\title{Nonparametric Structure Regularization Machine for 2D Hand Pose Estimation}

\newcommand*\samethanks[1][\value{footnote}]{\footnotemark[#1]}
\author[1]{Yifei Chen\thanks{These authors contribute equally to this work.}}
\author[2]{Haoyu Ma\samethanks}
\author[2]{Deying Kong}
\author[2]{Xiangyi Yan}
\author[1]{Jianbao Wu}
\author[1]{Wei Fan\thanks{These authors are co-corresponding authors of this work.}}
\author[2]{Xiaohui Xie\samethanks}
\affil[1]{Tencent Hippocrates Research Lab}
\affil[2]{Department of Computer Science, University of California at Irvine}
\affil[ ]{{\tt\small \{dolphinchen,jannwu,davidwfan\}@tencent.com, \{haoyum3,deyingk,x.yan,xhx\}@uci.edu}}

\maketitle
\ifwacvfinal\thispagestyle{empty}\fi

\begin{abstract}
Hand pose estimation is more challenging than body pose estimation due to severe articulation, self-occlusion and high dexterity of the hand.
Current approaches often rely on a popular body pose algorithm, such as the Convolutional Pose Machine (CPM), to learn 2D keypoint features.
These algorithms cannot adequately address the unique challenges of hand pose estimation, because they are trained solely based on keypoint positions without seeking to explicitly model structural relationship between them.
We propose a novel Nonparametric Structure Regularization Machine (NSRM) for 2D hand pose estimation, adopting a cascade multi-task architecture to learn hand structure and keypoint representations jointly.
The structure learning is guided by synthetic hand mask representations, which are directly computed from keypoint positions,
and is further strengthened by a novel probabilistic representation of hand limbs and an anatomically inspired composition strategy of mask synthesis.
We conduct extensive studies on two public datasets - OneHand 10k and CMU Panoptic Hand.
Experimental results demonstrate that explicitly enforcing structure learning consistently improves pose estimation accuracy of CPM baseline models, by 1.17\% on the first dataset and 4.01\% on the second one.
The implementation and experiment code is freely available online\footnote{\url{https://github.com/HowieMa/NSRMhand}}. 
Our proposal of incorporating structural learning to hand pose estimation requires no additional training information, and can be a generic add-on module to other pose estimation models. 


\end{abstract}

\section{Introduction}
Hand pose understanding is an important task for many real world AI applications, such as human-computer interaction, augmented reality and virtual reality.
However, hand pose estimation remains challenging because the hand is highly articulated and dexterous, and suffers severely from self-occlusion.
Recently a significant amount of efforts have been dedicated to improving the accuracy of hand pose estimation from different perspectives, including 1) multi-view RGB systems \cite{simon2017hand,joo2019panoptic}, 2) depth-based solutions \cite{ge2018hand,wan2018dense,yuan2018depth}, and 3) monocular RGB solutions \cite{zimmermann2017learning,panteleris2018using,cai2018weakly}. Although some of these efforts focus on 3D hand pose or shape estimation, 2D hand pose estimation remains an essential component as it often constitutes a sub-module of 3D estimation problems and as such directly impacts the performance of downstream 3D pose or shape estimation.

Meanwhile, human pose estimation has advanced significantly since the advent of {\em Deep Convolutional Neural Network} (DCNN). 
Successful DCNN architectures typically have large receptive fields and strong representation power, such as the {\em Convolutional Pose Machine} (CPM) \cite{wei2016convolutional}, the {\em Stacked Hourglass} (SHG) \cite{newell2016stacked}, and the {\em Residual Network} \cite{he2016deep}.
They are deployed by popular human pose estimation systems \cite{cao2018openpose,fang2017rmpe,he2017mask} to implicitly capture structure information of body parts.
They are also utilized by many hand pose estimation algorithms to perform the 2D pose estimation subtask \cite{simon2017hand,zimmermann2017learning,cai2018weakly,panteleris2018using,wan2018dense}.
However, DCNNs only capture structure information implicitly and may not be adequately equipped to capture complex structure relationship between hand keypoints to handle severe articulation and self-occlusion of the hand \cite{ke2018aware,kong2019adaptive}.

Recently, there is a trend to unify pose estimation and instance segmentation in a multi-task learning paradigm, and it is observed that the latter helps to improve the performance of the former \cite{he2017mask,wang2018mask}.
Unfortunately, this direction requires a large amount of manually labelled segmentation masks, which is costly to obtain.
Hand mask datasets are even rarer than body mask datasets, making the multi-task approach less applicable to hand pose estimation.

In this paper, we propose the Nonparametric Structure Regularization Machine (NSRM) for 2D hand pose estimation from a monocular RGB image.
NSRM incorporates a nonparametric structure model and a pose model in a cascade multi-task framework.
The structure learning is supervised by synthetic hand mask representations directly computed from keypoint positions, and is strengthened by a  probabilistic representation of hand limbs and an anatomically inspired composition strategy of mask synthesis.
The pose model utilizes the composite structure representation to learn robust hand pose representation.
We comprehensively evaluate the performance of NSRM on two public datasets, \ie, OneHand 10k \cite{wang2018mask}, and the more challenging CMU Panoptic Hand \cite{joo2019panoptic}.
Quantitative results demonstrate that NSRM consistently improves the prediction accuracy of the CPM baseline, by {\em 1.17\%} on the first dataset and {\em 4.01\%} on the second one, and that NSRM renders competitive performance compared to utilizing manually-labeled masks.
Qualitative results show that NSRM effectively reinforces structure consistency to predicted hand pose especially when severe occlusion exists, and the learned structure representations highly resemble real segmentation masks.

The main contributions of this paper are as follows:
\begin{itemize}
\item We propose a novel cascade structure regularization methodology for 2D hand pose estimation, which utilizes synthetic hand masks to guide keypoints structure learning. The synthetic masks are derived directly from keypoint positions requiring no extra data annotations, making the method applicable to any existing pose estimation model.  
\item We propose a novel probabilistic representation of hand limbs and an anatomically inspired composition strategy for hand mask synthesis.
\item We carry out extensive experiments on two public datasets, and demonstrate that NSRM consistently outperforms baseline models.  
\end{itemize}


\section{Related work}
\subsection{Human pose estimation}
DCNN has been massively applied to 2D human pose estimation since the seminal work of {\em DeepPose} \cite{toshev2014deeppose}. 
As the human body naturally manifests an articulated graph structure, researchers have explored the combination of DCNN and graphical models (GM) for pose estimation \cite{tompson2014joint,chen2014articulated,yang2016end,song2017thin}.
However, the GM component often suffers from two practical limitations: 
1) the pairwise term typically takes some parametric form, which may not be true in reality; 
2) belief propagation inference is performed frequently during training and computational intensive.  
As a result, mainstream algorithms \cite{cao2018openpose,fang2017rmpe,he2017mask} still rely on delicate DCNN architectures, such as CPM \cite{wei2016convolutional}, SHG \cite{newell2016stacked}, and the Residual Network \cite{he2016deep} to implicitly capture structure information, deploying their large receptive fields and strong representation power.
To further improve the effect of structure regularization, some approaches attempt to modify the output of DCNN, for example, to introduce extra branches of the {\em offset field} \cite{papandreou2017towards} or the {\em structural-aware loss} \cite{ke2018aware}.
Unfortunately, they still cannot fully characterize the structure conditions of limbs, especially their poses and interactions, resulting in very limited effect.
 
Meanwhile, 3D human pose estimation from monocular RGB, a challenger problem due to depth ambiguity, also advances significantly.
Some researches explicitly infer 3D coordinates from the 2D pose \cite{bogo2016keep,lee2018propagating,rayat2018exploiting,pavllo20193d}, while others incorporate 2D pose estimation networks into the whole architecture \cite{lin2017recurrent,mehta2017vnect,zhou2017towards}.
In both cases, DCNN-based 2D pose estimators are intensively utilized, such as the Mask RCNN \cite{he2017mask} and SHG.
To enforce structure constraints in 3D, a kinematic layer can be added on top of the network \cite{zhou2016deep}.
But this relies on known bone length and may suffer from the optimization difficulty, which limit its practical application.

\subsection{Hand pose estimation}
Hand pose estimation is more difficult than body pose estimation, as the hand is highly articulated, dexterous, and suffers severely from self-occlusion.
Although multi-view camera systems can solve the task reliably \cite{simon2017hand,joo2019panoptic}, it is usually very costly to build such a system, involving numerous optical devices and complicated configuration.
Therefore, their applications are mostly restricted to the laboratory scenarios.
To circumvent this limitation, researchers also devote much effort to depth-based solutions \cite{ge2018hand,wan2018dense,yuan2018depth}.
However, depth devices have limited resolution and range, and are sensitive to lighting conditions \cite{mehta2017vnect}.
And after all, they are still less ubiquitous than RGB cameras.

Due to the drawbacks of multi-view and depth-based solutions, monocular RGB approaches are drawing much more attention in recent years.
Like in the case of 3D human pose estimation, many algorithms adopt a {\em two-stage} framework, i.e., first performing 2D hand pose estimation and then lifting from 2D to 3D \cite{zimmermann2017learning,panteleris2018using,cai2018weakly}.
The 2D subtasks commonly utilize prevalent 2D human pose algorithms, in particular CPM and SHG, which are also frequently used in the multi-view or depth-based solutions \cite{simon2017hand,wan2018dense}.
Given the critical role of 2D hand pose algorithms in solving the complete 3D task, we focus on imposing novel structure regularization to 2D hand pose estimation in this paper.

\begin{figure*}[ht!]
\begin{center}
    \begin{subfigure}{.215\textwidth}
        \begin{center}
            \includegraphics[width=0.98\linewidth]{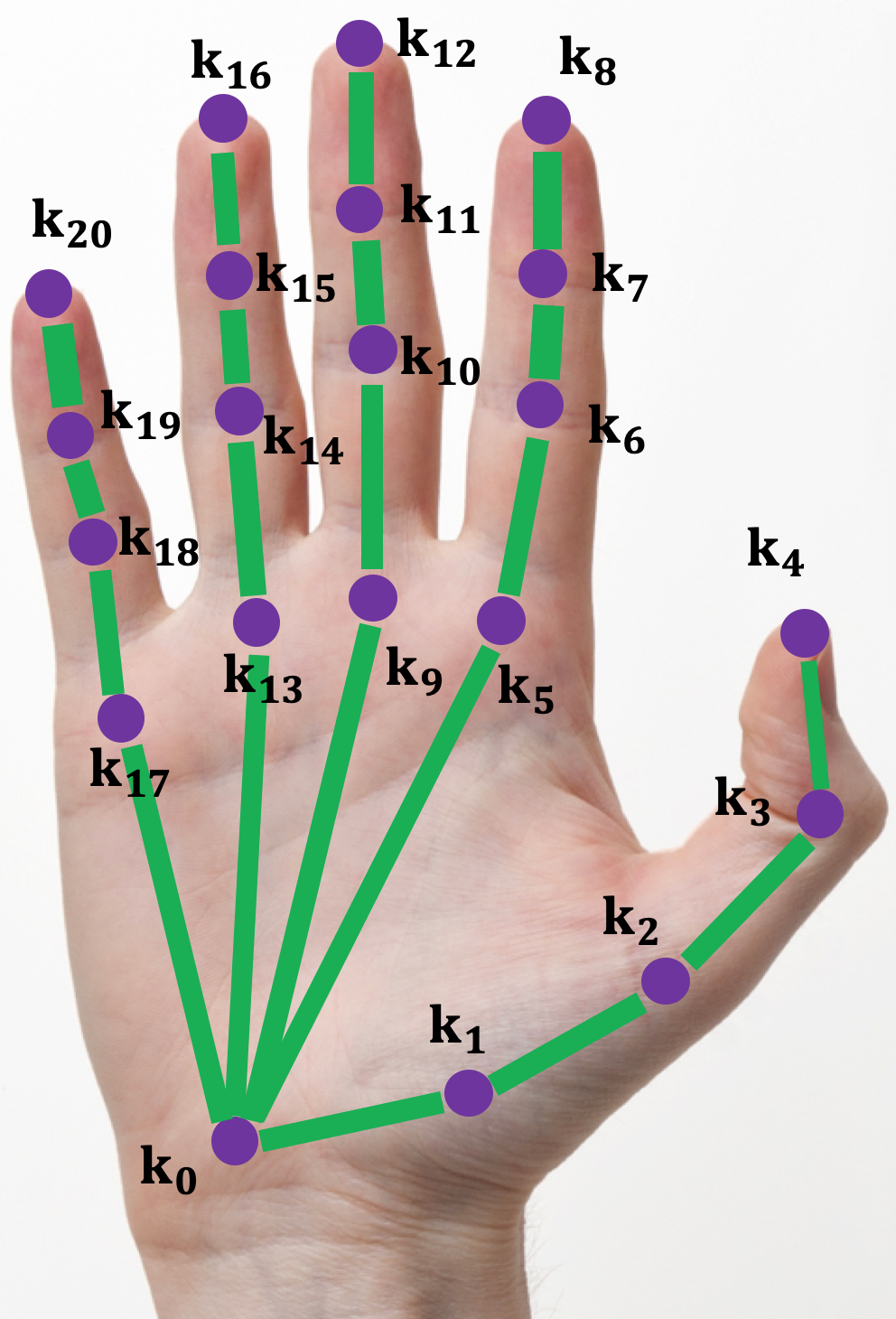}
        \end{center}
        \caption{the hand model}
        \label{fig:framework:hand}
    \end{subfigure}
    \begin{subfigure}{.639 \textwidth}
        \begin{center}
            \includegraphics[width=0.98\linewidth]{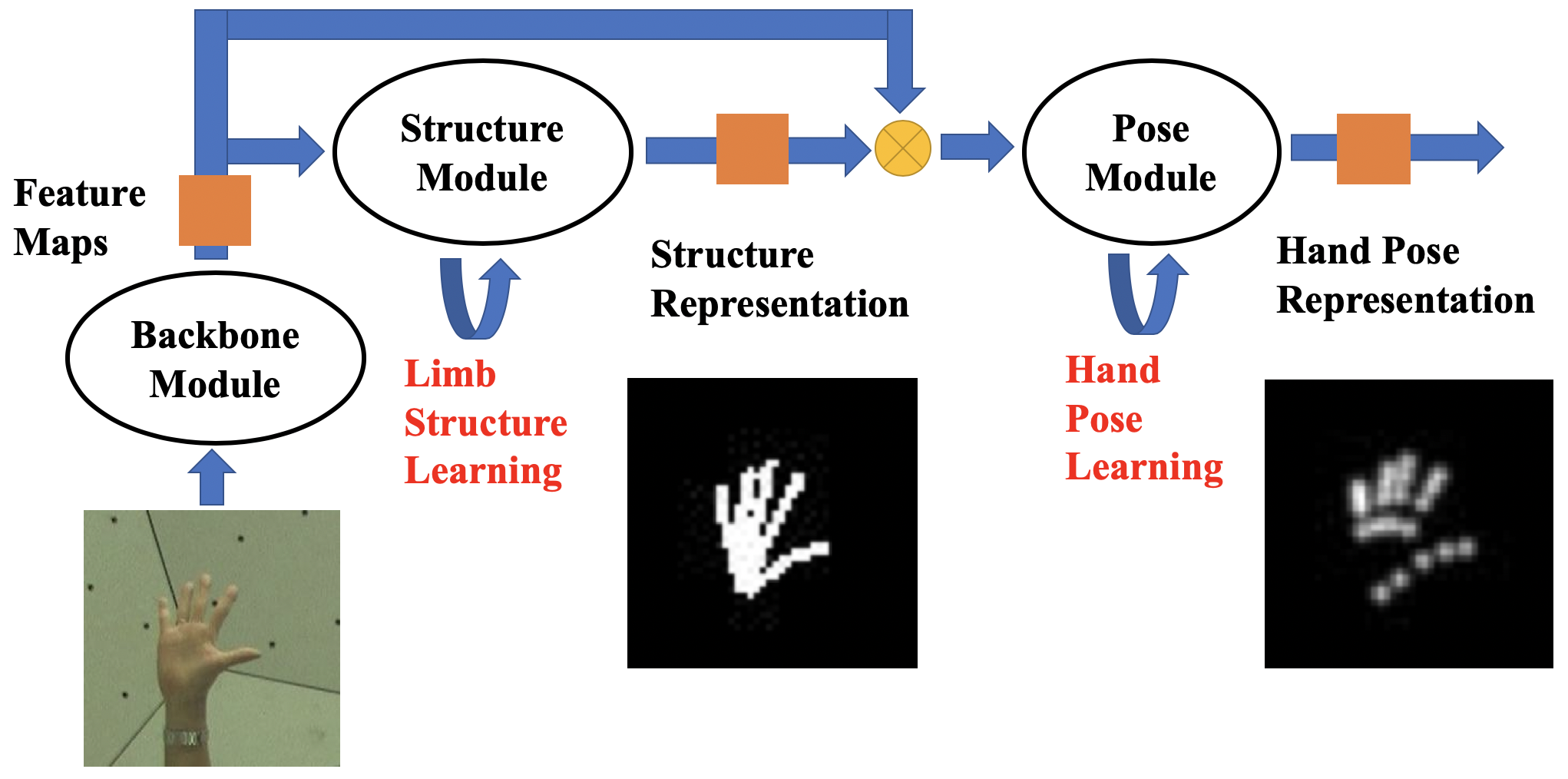}
        \end{center}
        \caption{the high-level architecture of NSRM}
        \label{fig:framework:model}
    \end{subfigure}
\caption{Illustration of the problem setting and our proposed framework. The hand is modeled as 21 keypoints, and 20 limbs that interconnect them anatomically. NSRM features a hierarchical multi-task architecture that learns structure representation and keypoint representation sequentially, which is generic for multi-stage pose estimation models such as CPM \cite{wei2016convolutional} and SHG \cite{newell2016stacked}. Structure learning is guided by our novel synthetic hand mask representations (see \ref{sec:limb:repr}, \ref{sec:limb:group} for details).}
\label{fig:framework}
\end{center}
\end{figure*}

\section{The model}
Nonparametric Structure Regularization Machine (NSRM) learns the hand's structure and keypoint representations in a cascade multi-task framework.  
A high-level illustration of the hand representation and our overall architecture is shown in Figure~\ref{fig:framework}. 
The hand is modeled as 21 keypoints and 20 limbs.
The former are interconnected via the latter anatomically. 
First, the backbone module, usually a DCNN, \eg, the VGG \cite{simonyan2015vgg}, processes an input image and generates a set of feature maps. 
Second, the structure module takes in image feature maps to learn the mask representation of the hand. 
Third, the pose module takes in both the mask representation and the feature maps to learn the pose representation of the hand, i.e., {\em keypoint confidence maps} (KCM). 
Both the structure and the pose modules are multi-stage convolutional neural networks.

\subsection{Limb mask representation}
\label{sec:limb:repr}
\begin{figure}[h!]
\begin{center}
    \begin{subfigure}{.43 \linewidth}
        \begin{center}
             \includegraphics[width=0.98\textwidth]{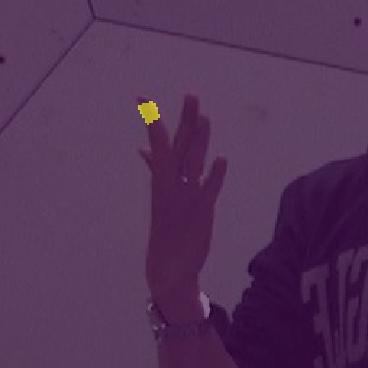}
        \end{center}
        \caption{LDM}
        \label{fig:repr:lam}
    \end{subfigure}
    \begin{subfigure}{.43 \linewidth}
        \begin{center}
            \includegraphics[width=0.98\textwidth]{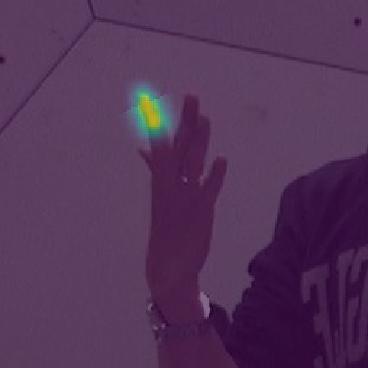}
        \end{center}
        \caption{LPM}
        \label{fig:repr:lsh}
    \end{subfigure}
\end{center}
\caption{Two limb mask representations (taking the index finger tip as an example). LDM: Limb Deterministic Mask (Equation~\ref{eq:def:ldm}); LPM, Limb Probabilistic Mask (Equation~\ref{eq:def:lpm}).}
\label{fig:repr}
\end{figure}

Consider any particular limb $L$ between Keypoint $i$ and $j$ as defined in Figure \ref{fig:framework:hand}.
We define our basic mask representation as follows,

{\bf Limb Deterministic Mask (LDM)}.
Pixels that belong to $L$ are defined as those which fall inside a fixed-width rectangle centering around line segment $\overline{p_i p_j}$, \ie,
\begin{align}
\begin{cases}
   0\leq \left(p - p_j \right)^T \left(p_i - p_j\right) \leq \| p_i - p_j \|_2^2, \\
   \left| \left(p - p_j \right)^T \mathbf{u}^\perp \right| \leq \sigma_{LDM}
\end{cases}
\end{align}
where $\mathbf{u}^\perp$ is a unit vector perpendicular to $\overline{p_i p_j}$, and $\sigma_{LDM}$ is a hyper parameter to control the width of the limb.
The ground truth of LDM is a simple 0/1-mask defined outside/inside the rectangle, \ie, 
\begin{align}
\label{eq:def:ldm}
S_{LDM}\left( p|L \right) = \begin{cases}
                       1 & \text{if} \; p \in L\\
                       0 & \text{otherwise}
               \end{cases}
\end{align}
where $p\in I$ is an arbitrary pixel in the image.
See Figure \ref{fig:repr:lam} for an illustration.

{\bf Limb Probabilistic Mask (LPM)}.
LDM assigns 0/1 value to each pixel, depending on its belong to the rectangular mask.
This crude treatment may not be optimal in practice.
We further propose the novel LPM representation.
Each pixel belongs to $L$ with a Gaussian-alike confidence value as defined bellow,
\begin{align}
\label{eq:def:lpm}
S_{LPM}\left( p|L \right) = \exp \left( -\frac{\mathcal{D}(p, \overline{p_i p_j})}{2\sigma^2_{LPM}} \right)
\end{align}
where $\mathcal{D}(p, \overline{p_i p_j})$ is the distance between the pixel $p$ and the line segment $\overline{p_i p_j}$, and $\sigma_{LPM}$ is a hyper parameter to control the spread of the Gaussian.
See Figure~\ref{fig:repr:lsh} for an illustration.
LPM is a smoothed expansion of LDM.

\subsection{Limb composition}
\label{sec:limb:group}
\begin{figure}[h!]
\begin{center}
    \begin{subfigure}{.43 \linewidth}
        \begin{center}
             \includegraphics[width=0.98\textwidth]{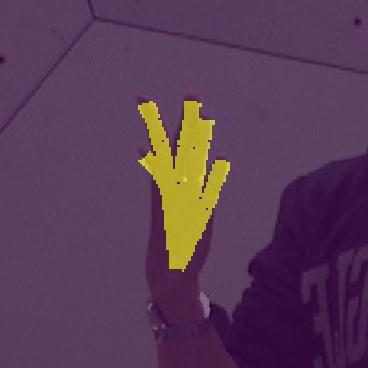}
        \end{center}
        \caption{LDM-G1}
        \label{fig:g1:lam}
    \end{subfigure}
    \begin{subfigure}{.43 \linewidth}
        \begin{center}
            \includegraphics[width=0.98\textwidth]{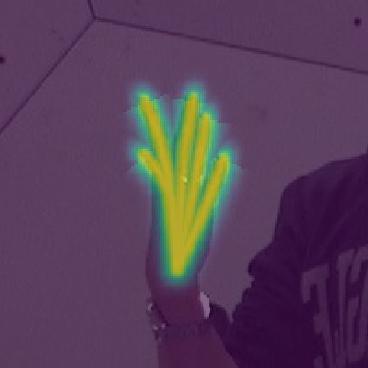}
        \end{center}
        \caption{LPM-G1}
        \label{fig:g1:lsh}
    \end{subfigure}
\end{center}
\caption{The G1 composition strategy: coalescing all the 20 limbs together to get one single mask representing the whole hand.}
\label{fig:g1}
\end{figure}

\begin{figure}[h!]
\begin{center}
	\includegraphics[width=0.95 \linewidth]{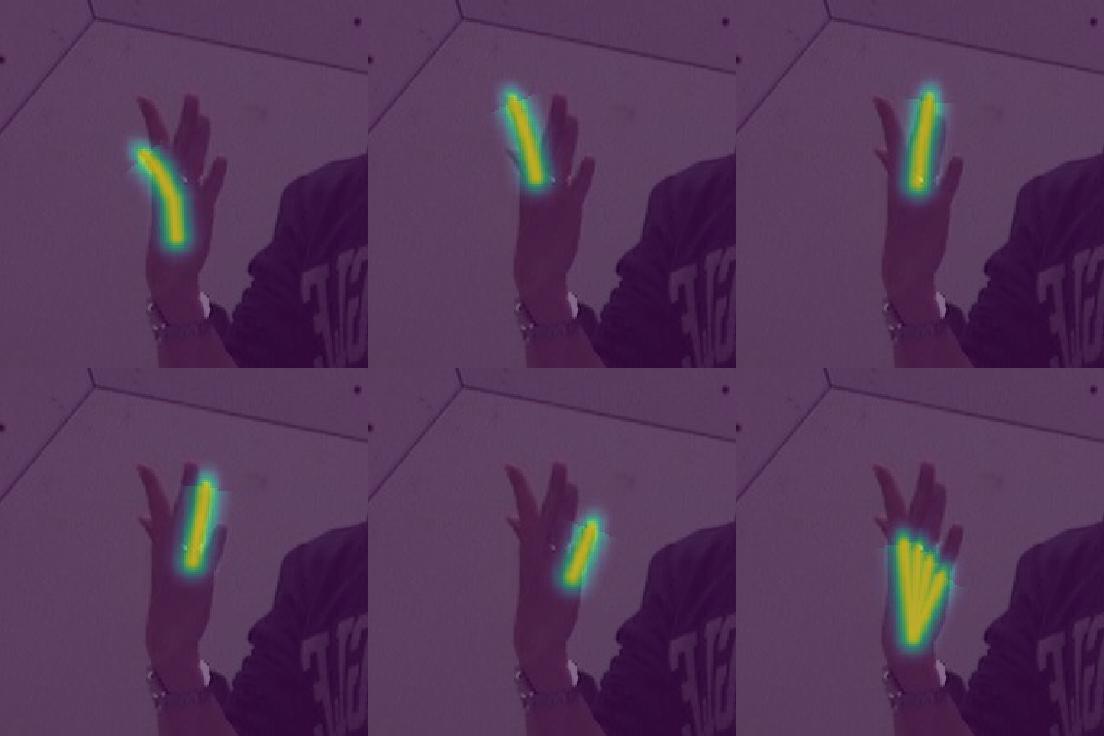}
\end{center}
\caption{The G6 composition strategy (using LPM representation, \eg, LPM-G6): coalescing the 20 limbs into 6 groups, representing five fingers and the palm separately.}
\label{fig:g6}
\end{figure}

Given mask representations of single limbs, we further coalesce them into anatomically legitimate groups.
Our basic strategy is to coalesce all the 20 limbs together, which renders one single mask representing the whole hand (denoted as {\em G1}).
Alternatively, we also consider coalescing limbs separately into six groups, one representing each finger and one representing the palm (denoted as {\em G6}).
G1 captures the overall structure of the hand, while G6 concerns more about detailed structure regarding local areas of the hand.
See Figure~\ref{fig:g1} and~\ref{fig:g6} for an illustration.
Formally, consider any particular limb group $g$ containing $|g|$ limbs, $\{L_1, L_2, ..., L_{|g|}\}$.
Using limb composition, the coalesced mask is defined as,
\begin{align}
\label{eq::cpst}
S^*\left( p | g \right) = \max \left( S(p|L_1), S(p|L_2), ..., S(p|L_{|g|}) \right) 
\end{align}
where $S(p|L)$ is computed using the basic representation of $S_{LDM}$ (Equation~\ref{eq:def:ldm}) or $S_{LPM}$ (Equation \ref{eq:def:lpm}).

In practice, we mainly focus on utilizing G1 and {\em G1\&6} (the combination of G1 and G6).
We note that although G1 resembles the hand segmentation mask, it is much more efficient because it is readily obtained from keypoins without the extra work of mask annotation.
In Section~\ref{sec:exp}, we will compare the performance of utilizing our LDM/LPM representations against the real segmentation mask.

\subsection{Loss function and training}
Intermediate supervision is applied to each stage of the structure module and the pose module.
Following the common practice of instance segmentation \cite{he2017mask}, we apply cross-entropy loss to the output of our structure module, \ie,
\begin{align}
\mathcal{L}_S = \sum_{t=1}^{T_S} \sum_{g \in G} & \sum_{p\in I} S^*(p | g)\log\hat{S}_t (p | g) \\ \nonumber
                                                                + & \left(1 - S^*(p | g)\right) \log \left(1 - \hat{S}_t (p | g)\right)
\end{align}
where $T_S$ is the number of stages of structure learning, and $\hat{S}_t(p|g)$ is the prediction of the structure module at Pixel $p$, Group $g$ of Stage $t$.

Following the common practice of pose estimation \cite{newell2016stacked,cao2018openpose}, we define the ground truth KCM of Keypoint $k$ as a 2D Gaussian centering around the labelled keypoint with standard deviation $\sigma_{KCM}$, \ie,
\begin{align}
C^*(p|k) = \exp \left\{ -\frac{\| p-p^*_k \|_2^2}{2 \sigma_{KCM}^2} \right\}
\end{align}
We apply the sum-of-squared-error loss to the output of our pose module, \ie,
\begin{align}
\mathcal{L}_K = \sum_{t=1}^{T_K} \sum_{k=1}^{K} \sum_{p \in I} \left\| C^*(p | k) -\hat{C}_t(p | k) \right\|_2^2
\end{align}
where $T_K$ is the number of stages of pose learning, and $\hat{C}_t (p|k)$ is the prediction of the pose module at Pixel $p$, Keypoint $k$ of Stage $t$.

The overall loss function is thus a weighted sum of the structure loss and the pose loss, \ie,
\begin{align}
\label{eq::loss}
\mathcal{L} = 
\begin{cases}
	\mathcal{L}_{K} + \lambda_1 \mathcal{L}_S^{G1}, & \text{for G1} \\
	\mathcal{L}_{K} + \lambda_1 \mathcal{L}_S^{G1} + \lambda_2 \mathcal{L}_S^{G6}, & \text{for G1\&6}
\end{cases}
\end{align}
where $\lambda_1, \lambda_2$ are hyper-parameters to control the relative weight of the structural regularization.
The whole system is trained end-to-end.

\section{Experiments}
\label{sec:exp}

\subsection{Datasets}
We evaluate NSRM on two public hand pose datasets: the OneHand 10k dataset \cite{wang2018mask} (OneHand 10k), and the CMU Panoptic Hand dataset \cite{joo2019panoptic} (Panoptic).
Their overall statistics are summarized in Table \ref{tab:data:stat}.
More descriptions are as follows.

{\bf OneHand 10k} contains 11,703 in-the-wild hand images annotated with both segmentation masks and keypoints.
Being collected online, the images often have cluttered background and cover various hand poses.
Invisible keypoints are often left unannotated.
Ground truth limb representations related to these missing keypoints are set to zero maps.
We don't do any hand cropping, as most of the images are occupied by one hand.
The dataset is already partitioned into training and testing subsets by Wang \etal \cite{wang2018mask}.
 
{\bf Panoptic} contains 14,817 images of persons from the Panoptic Studio, each with 21 annotated keypoints of the right hand.
Since we focus on hand pose estimation instead of hand detection, we directly crop hands based upon their ground truth keypoints.
Specifically, we crop a square patch of size 2.2B, where B is the maximum dimension of the tightest bounding box enclosing all hand keypoints. 
Then the cropped hand dataset is randomly divided into three subsets for training (80\%), validation (10\%) and testing (10\%).

\begin{table}[h!]
\caption{Overall statistics of datasets used in this paper.}
\label{tab:data:stat}
\begin{center}
    \begin{tabular}{llll}
    \toprule
    dataset & training & validation & testing \\
    \midrule
    OneHand 10k & 10,000 & - & 1,703 \\
    Panoptic & 11,853 & 1,482 & 1,482 \\
    \bottomrule
    \end{tabular}
\end{center}
\end{table}

\subsection{Experimental settings}
\noindent {\bf Implementation details}

\begin{figure*}[t!]
	\begin{center}
		\includegraphics[width=0.9 \linewidth]{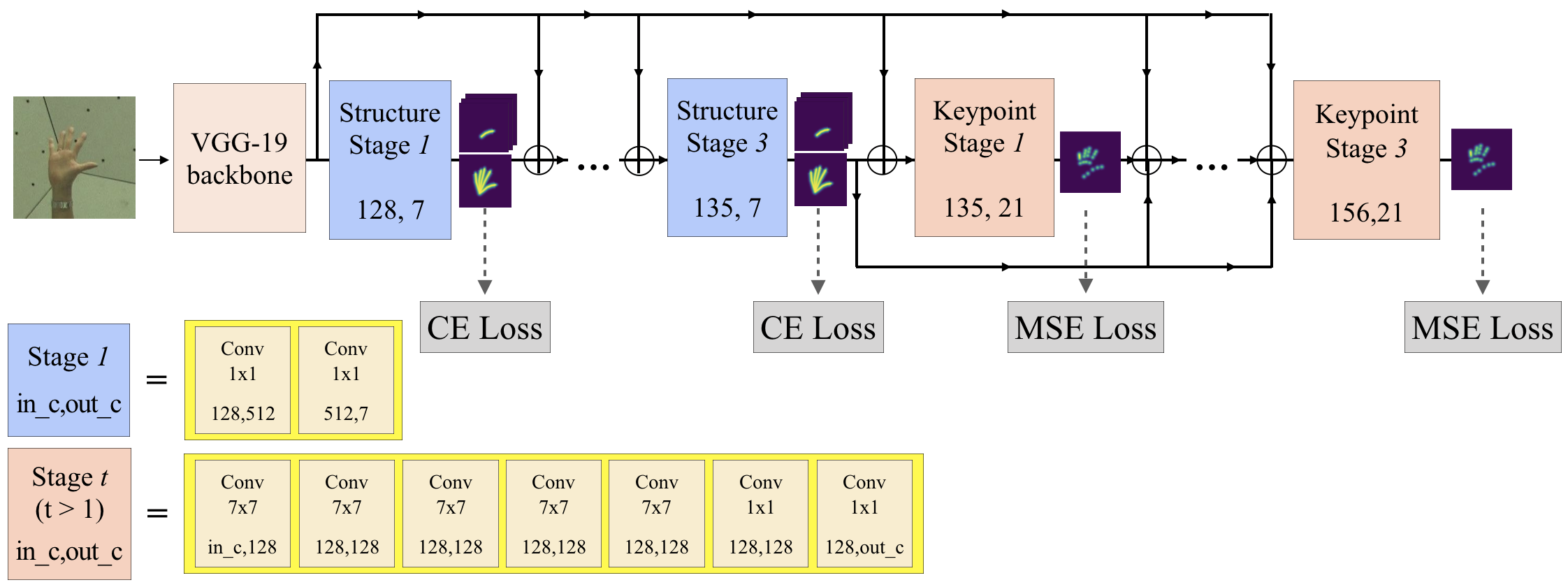}
	\end{center}
        \caption{Detailed architecture of NSRM using G1\&6 composition implemented based upon CPM.
        Each stage consists of a series of fully convolutional layers, whose kernel size is denoted as $k\times k$.
        The number of input feature maps and the number of output feature maps are shown lower insider each rectangle.
        The output of Structure Stage 3 is fed to each keypoint stage.}
        \label{fig:detail_arc}
\end{figure*}

We implement NSRM in Pytorch.
To be compatible with and comparable to the CPM hand pose model \cite{simon2017hand}, we adopt VGG-19 \cite{simonyan2015vgg} (up to Conv 4\_4) as our backbone network, which is pretrained on ImageNet \cite{deng2009imagenet} and generates 128-channel feature maps.
The following architecture has 6 stages, each of which contains 5 convolution layers with 7x7 kernel and 2 convolution layers with 1x1 kernel (except the first stage).
The first 3 stages are allocated to learn composite mask representations, and the last 3 stages are for pose representation learning.
All hand image patches are resized to $368 \times 368$ before fed into our model, yielding $46 \times 46$ representation maps for both  LDM/LPM and KCM.
The detail network architecture is shown in Figure~\ref{fig:detail_arc}.  

\noindent {\bf Learning configuration}

We use Adam \cite{kingma2014adam} to train our model.
The initial learning rate is set to 1e-4, and other parameters are set to default values.
For G1, we set $ \lambda_1 = 1$ for LDM, and $ \lambda_1 = 0.5$ for LPM. 
For G1\&6, we set $\lambda_1 = 0.2, \lambda_2 = 0.04$ for LDM, and $\lambda_1 = 0.1, \lambda_2 = 0.02$ for LPM.
These configurations empirically make the structure loss and the pose loss on the same scale at the beginning of training.
Further more, as structure learning is an auxiliary task and our ultimate goal is hand pose estimation, we propose to utilize a {\em decayed loss training schedule}.
Specifically, we let $ \lambda_1$ and $ \lambda_2$ decay by a ratio of 0.1 every 20 epochs, so as to let training focus more on KCM in later iterations.



\noindent {\bf Evaluation metric}

We adopt {\em Probability of Correct Keypoint within a Normalized Distance Threshold} \cite{simon2017hand} (shortly referred to as PCK) to perform model selection and evaluation.
However, as the hand/head sizes are not explicitly provided by the datasets used in this paper, we resort to normalization with respect to the dimension of the tightest hand bounding box.
The normalization threshold $\sigma_{PCK}$ ranges from 0 to 1.

\subsection{Quantitative results}
\label{sec:exp:quant-rslt}

\noindent \textbf{OneHand 10k}

\begin{figure*}[h!]
\begin{center}
    \begin{subfigure}{.46 \linewidth}
        \begin{center}
            \includegraphics[width=0.98\textwidth]{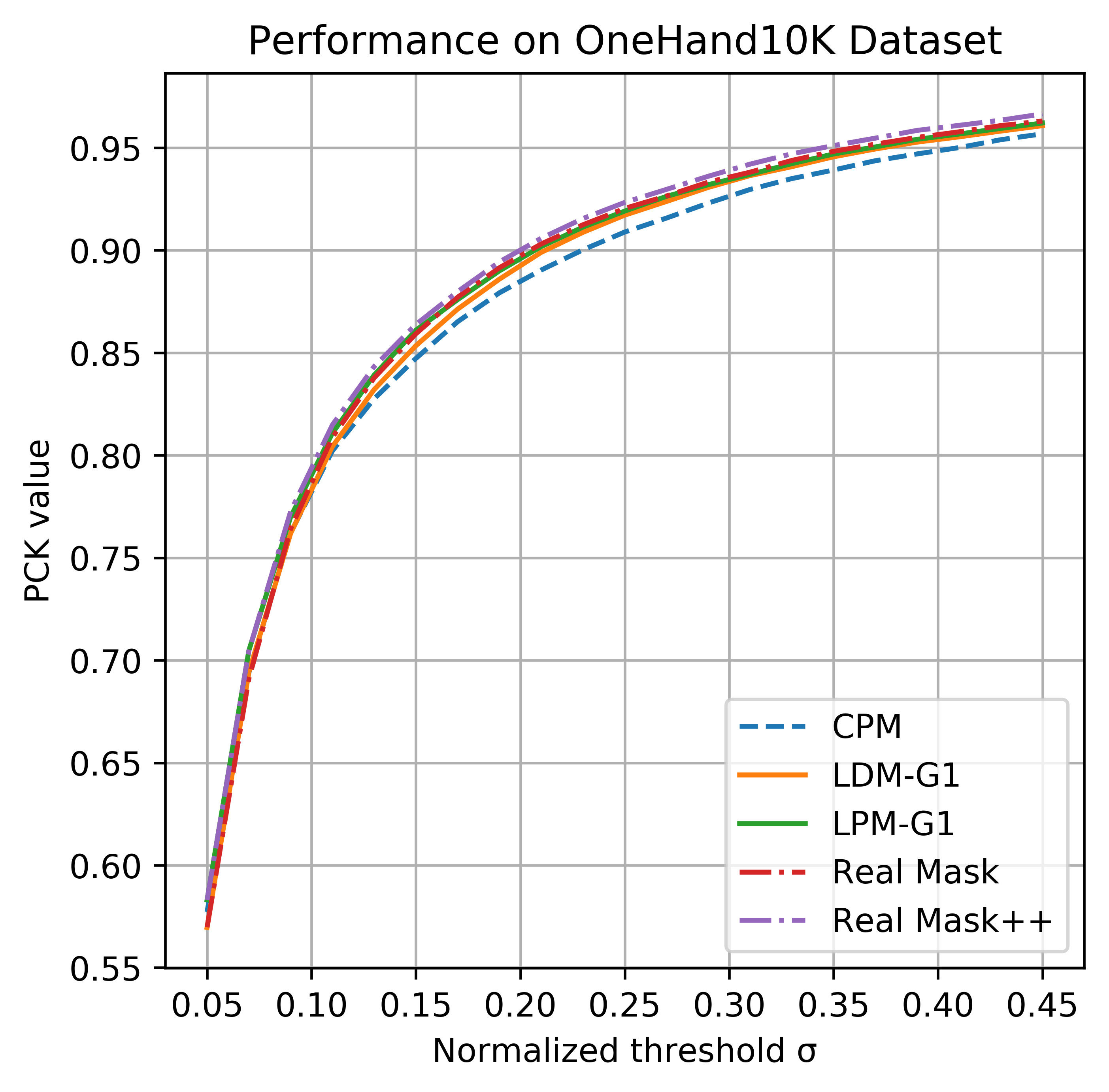}
        \end{center}
        \caption{}
        \label{fig:pck:onehand}
    \end{subfigure}
    \begin{subfigure}{.46 \linewidth}
        \begin{center}
             \includegraphics[width=0.98\textwidth]{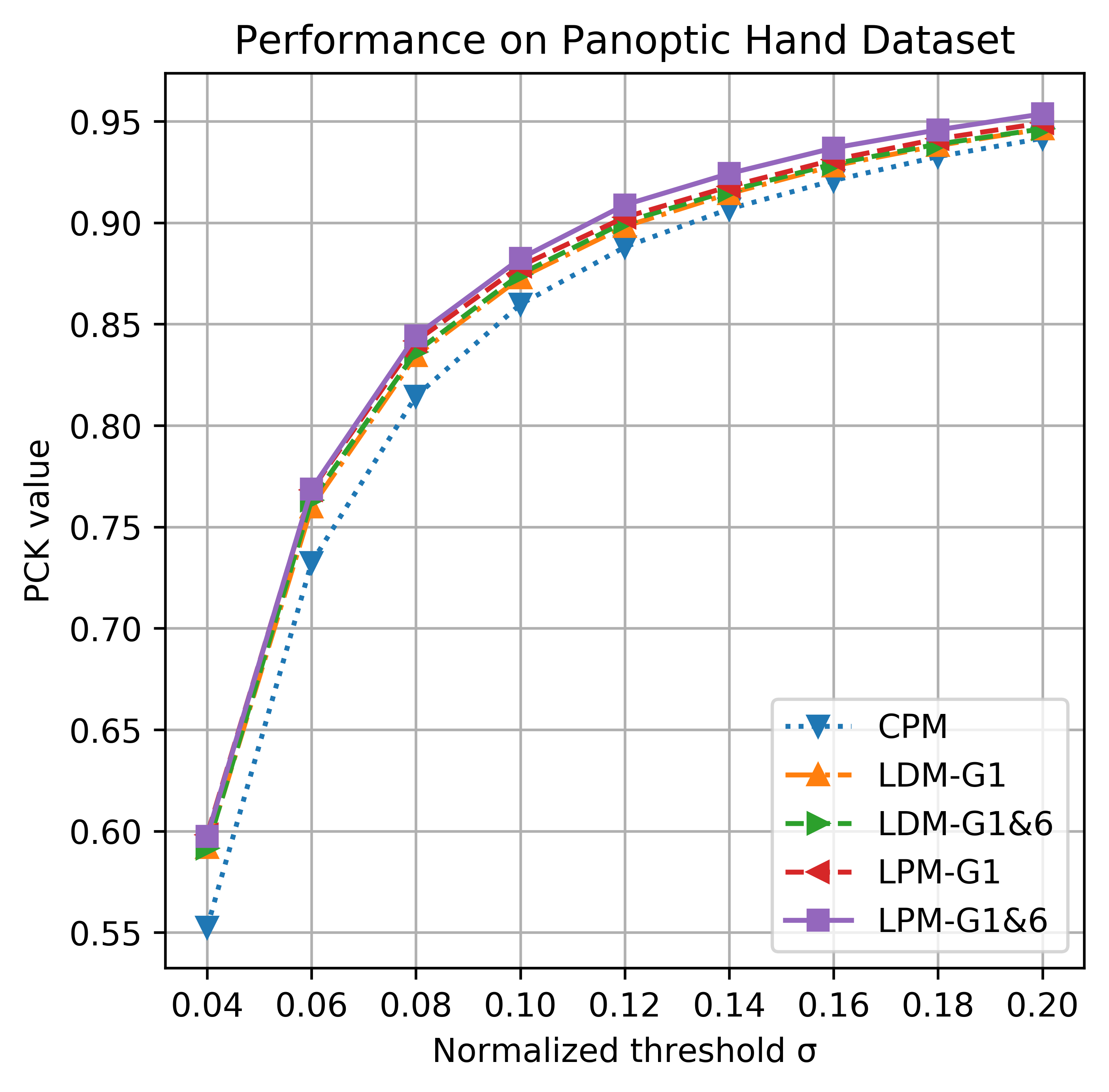}
        \end{center}
        \caption{}
        \label{fig:pck:panoptic}
    \end{subfigure}
\end{center}
\caption{PCK curves on testing data of a) OneHand 10k, and b) Panoptic. Best viewed in color.}
\label{fig:pck}
\end{figure*}

As this dataset has a lot of missing values, G6 composition tends to generate incomplete masks.
Therefore, we only consider about G1 composition.
We retrain the Mask-pose Cascaded CNN \cite{wang2018mask}, which utilize real segmentation masks (denoted as ``Real Mask").
We also train the model with our proposed decayed loss training schedule (denoted as ``Real Mask ++").
Figure \ref{fig:pck:onehand} shows the performance comparison.
Table \ref{tab:pck:onehand} summarizes detailed numerical results.
Our observations are as follows:

i) NSRM consistently improves the predictive accuracy of CPM, regardless of the choice of basic representation (LDM or LPM), and the value of the evaluation threshold $\sigma_{PCK}$.
In particular, LPM-G1 achieves {\em 0.0102} absolute improvement in average PCK, corresponding to {\em 1.17\%} relative improvement.

ii) LPM-G1 outperforms LDM-G1, and does slightly better than Real Mask.
This result demonstrates the effectiveness of our proposed probabilistic mask, comparing to both the manually labeled mask and our proposed deterministic mask.

iii) Our proposed decayed loss training schedule, in combination with utilizing real masks, achieves the best performance (Real Mask ++).

To summarize, our proposed NSRM (along with its learning schedule) is both efficient and effective, as it avoids the overhead of mask labeling but still maintains competitive performance.

\begin{table*}[h!]
    \caption{Detailed numerical results of PCK (in \%) evaluated at different thresholds on the OneHand 10k testing data. ``ave" means the average PCK, whose absolute and relative improvement are shown in the right most column. The best improvement is highlighted in boldface.}
    \label{tab:pck:onehand}
    \begin{center}
        \begin{tabular}{llllllll}
            \toprule
            $\sigma_{PCK}$ & 0.1 & 0.15 & 0.2 & 0.25 & 0.3 & ave & improvement \\
            \midrule
            CPM & 78.48 & 84.73 & 88.54 & 90.89 & 92.64 & 87.06 & - \\
            \midrule
            LDM-G1 & 78.50 & 85.35 & 89.31 & 91.72 & 93.35 & 87.64 & +0.59 (+0.67\%) \\
            LPM-G1 & 79.32 & 86.10 & 89.60 & 91.91 & 93.43 & 88.07 & +1.02 (+1.17\%) \\
            \midrule
            Real Mask \cite{wang2018mask} & 78.95 & 85.93 & 89.78 & 92.04 & 93.55 & 88.05 & +0.99 (+1.14\%) \\
            Real Mask ++ & 79.62 & 86.38 & 90.05 & 92.34 & 93.92 & 88.46 & {\bf +1.41 (+1.62\%)} \\
            \bottomrule
        \end{tabular}
    \end{center}
\end{table*}

\noindent {\bf Panoptic}

Figure \ref{fig:pck:panoptic} shows the performance of NSRM and the CPM baseline.
Table \ref{tab:pck:panoptic} summarizes detailed numerical results.
Our observations are as follows:

i) Like in the case of OneHand 10k, NSRM consistently outperforms the CPM baseline.
In particular, the fully-fledged LPM-G1\&6 achieves {\em 0.0309} absolute improvement in average PCK, corresponding to {\em 4.01\%} relative improvement.
The results suggest a systematic and significant benefit of utilizing our proposed structure regularization for 2D hand pose estimation.

ii) Like in the case of OneHand 10k, LPM consistently outperforms LDM, under both composition strategies (G1 and G1\&6). 
The absolute improvement in average PCK is 0.64 under G1 and 0.71 under G1\&6.
This phenomenon consolidates that the probabilistic treatment in mask representation indeed benefits the performance of NSRM.

iii) Comparing the proposed composition strategies, we find that G1\&6 moderately improves the performance of G1.
We interpret this result from an anatomical perspective.
As an overall representation, G1 covers important global structure information of the hand.
However it cannot fully characterize local details, such as the shape of each finger, which are highly flexible and hard to distinguish due to self-occlusion. 
G6 is designed to cope with this problem.
By combining G1 and G6, NSRM gets a representation that can cover both global and local structure information.   

iv) The combination of centralized \& distributed composition and the probabilistic representation (LPM-G1\&6) renders the optimal performance, which correspond to 0.89 absolute improvement in average PCK, comparing to the basic version (LDM-G1).

v) Although being consistent on both datasets, the improvement seems much more significant on Panoptic than on OneHand 10k.
There are two potential reasons.
First, Panoptic is much more challenging, with abundant hand gestures shot from different perspectives and complicated hand-hand \& hand-object interactions.
Second, OneHand 10K contains a lot of unlabelled keypoints, and therefore tends to make the mask learning signal too fragmented and noisy.
Both factors suggest that Panoptic could benefit more from structure learning than OneHand 10k.

\begin{table*}[h!]
\caption{Detailed numerical results of PCK (in \%) evaluated at different thresholds on the Panoptic testing data. ``ave" means the average PCK, whose absolute and relative improvement are shown in the right most column. The best improvement is highlighted in boldface.}
\centering
\begin{tabular}{llllllll}
\toprule
$\sigma_{PCK}$ & 0.04 & 0.06 & 0.08 & 0.10 & 0.12 & ave & improvement \\
\midrule
CPM & 55.25 & 73.23 & 81.45 & 85.97 & 88.80 & 76.94 & - \\
\midrule
LDM-G1 & 59.20 & 75.98 & 83.45 & 87.28 & 89.81 & 79.14 & +2.20 (+2.86\%) \\
LDM-G1\&6 & 59.16 & 76.32 & 83.63 & 87.46 & 90.03 & 79.32 & +2.38 (+3.09\%) \\
\midrule
LPM-G1 & 59.81 & 76.82 & 84.16 & 87.86 & 90.26 & 79.78 & +2.84 (+3.69\%) \\
LPM-G1\&6 & 59.73 & 76.86 & 84.43 & 88.23 & 90.87 & 80.03 & {\bf +3.09 (+4.01\%)} \\
\bottomrule
\end{tabular}
\label{tab:pck:panoptic}
\end{table*}

\subsection{Qualitative results}
\begin{figure*}[h!]
    \begin{center}
        \begin{subfigure}{.155 \textwidth}
            \begin{center}
                \includegraphics[width=0.98\textwidth]{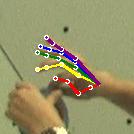}
            \end{center}
        \end{subfigure}
        \begin{subfigure}{.155 \textwidth}
            \begin{center}
                \includegraphics[width=0.98\textwidth]{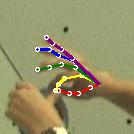}
            \end{center}
        \end{subfigure}
        \begin{subfigure}{.155 \textwidth}
            \begin{center}
                \includegraphics[width=0.98\textwidth]{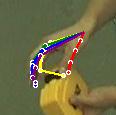}
            \end{center}
        \end{subfigure}
        \begin{subfigure}{.155 \textwidth}
            \begin{center}
                \includegraphics[width=0.98\textwidth]{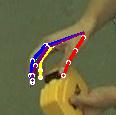}
            \end{center}
        \end{subfigure}
        \begin{subfigure}{.155 \textwidth}
            \begin{center}
                \includegraphics[width=0.98\textwidth]{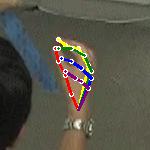}
            \end{center}
        \end{subfigure}
        \begin{subfigure}{.155 \textwidth}
            \begin{center}
                \includegraphics[width=0.98\textwidth]{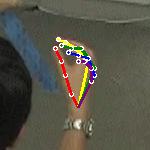}
            \end{center}
        \end{subfigure}
        \begin{subfigure}{.155 \textwidth}
            \begin{center}
                \includegraphics[width=0.98\textwidth]{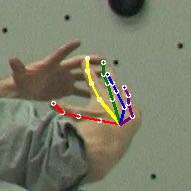}
            \end{center}
        \end{subfigure}
        \begin{subfigure}{.155 \textwidth}
            \begin{center}
                \includegraphics[width=0.98\textwidth]{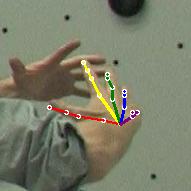}
            \end{center}
        \end{subfigure}
        \begin{subfigure}{.155 \textwidth}
            \begin{center}
                \includegraphics[width=0.98\textwidth]{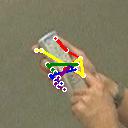}
            \end{center}
        \end{subfigure}
        \begin{subfigure}{.155 \textwidth}
            \begin{center}
                \includegraphics[width=0.98\textwidth]{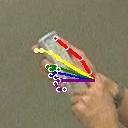}
            \end{center}
        \end{subfigure}
        \begin{subfigure}{.155 \textwidth}
            \begin{center}
                \includegraphics[width=0.98\textwidth]{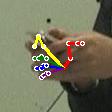}
            \end{center}
        \end{subfigure}
        \begin{subfigure}{.155 \textwidth}
            \begin{center}
                \includegraphics[width=0.98\textwidth]{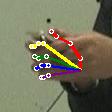}
            \end{center}
        \end{subfigure}
    \end{center}
    \caption{Visualization of predicted hand pose of samples from the Panoptic test data.
    For each pair of images, the left shows CPM's prediction and the right shows LPM-G1\&6's prediction.
    Best viewed in color.}
    \label{fig:quali:skl}
\end{figure*}

\begin{figure*}[h!]
    \begin{center}
        \includegraphics[width=0.92\textwidth]{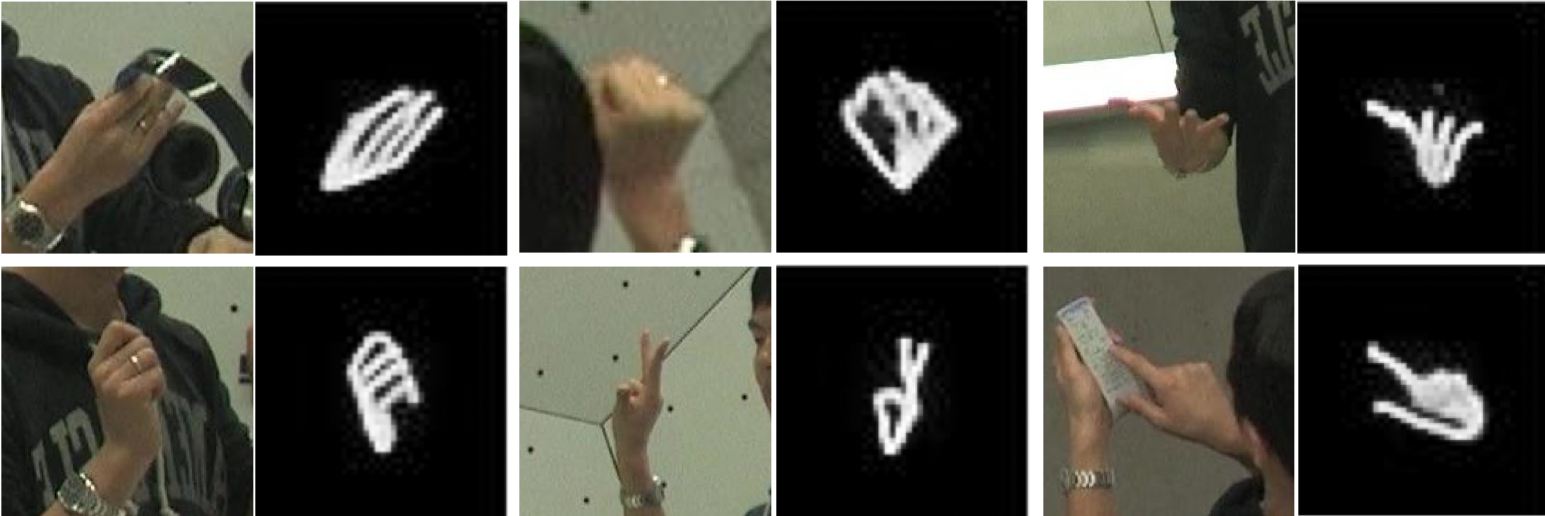}
    \end{center}
    \caption{Visualization of the learned structure representation of samples from the Panoptic test data.
    For each pair of images, the left shows the original image and the right shows the learned mask representation of LPM-G1, \ie, the output of Structure Stage 3 (see Figure~\ref{fig:detail_arc} for an illustration).}
    \label{fig:quali:mask}
\end{figure*}

Figure~\ref{fig:quali:skl} visualizes the predictive results of CPM and NSRM on sample images from the Panoptic test data.
We can clearly see that NSRM effectively reinforces structure consistency and reduces inference ambiguity.
Even during severe occlusion, LPM-G1\&6 still makes anatomically legitimate prediction while CPM cannot.
Moreover, Figure~\ref{fig:quali:mask} visualizes the learned structure representations on sample images.
We can see that they highly resemble hand segmentation masks.
This indicates that our NSRM framework could be potentially applied to multi-task learning of both hand pose estimation and instance segmentation.

\subsection{Discussion}
NSRM learns compositional structure representation of the hand, and utilizes it to regularize the learning process of KCM in a nonparametric fashion.
This implicit hierarchical treatment is fundamentally different to graphical-model-involved approaches \cite{chen2014articulated,yang2016end,song2017thin,kong2019adaptive}, and those which introduce simple keypoint-induced output/loss \cite{papandreou2017towards,ke2018aware} or kinematic constraints \cite{zhou2016deep}.

Our basic LDM representation takes a rectangular shape, similar to the Part Affinity Field (PAF) \cite{cao2018openpose}.
However, NSRM goes significantly beyond PAF in three core aspects.
First, our segmentation-inspired representation and its probabilistic expansion are completely different from the vector field representation of PAF.  
Second, we propose structure composition to coalescing limbs into anatomically inspired groups, a key feature not considered by PAF.
Last but not least, PAF is an auxiliary representation proposed for differentiating multiple human instances, not intended for hand structure regularization as in our case.

Previous researches have explored simultaneous pose estimation and instance segmentation \cite{he2017mask,wang2018mask}, but all require mask annotation.
Our structure representation is automatically constructed from keypoints, avoiding laborious annotation.
Our experiments demonstrate that NSRM achieves comparable pose estimation performance to models trained with real masks \cite{wang2018mask}.
More over, our learned structure representations highly resemble real masks, which indicates potential applications to hand instance segmentation.

\section{Conclusion}
In this paper, we have proposed a novel Nonparametric Structure Regularization Machine for 2D hand pose estimation.
NSRM is a cascade architecture of structure learning and pose learning.
The structure learning is guided by self-organized hand mask representations, and strengthened by a novel probabilistic representation of hand limbs and an anatomically inspired composition strategy of mask synthesis.
The pose module utilizes the structure representation to learn robust hand pose representation.
We comprehensively evaluate NSRM on two public datasets.
Experimental results demonstrate that,
1) NSRM consistently improves the prediction accuracy of the baseline model;
2) NSRM renders comparable performance to utilizing manually labeled masks;
3) NSRM effectively reinforces structure consistency to predicted hand poses, especially during severe occlusion.
We should note that although we used CPM as our baseline pose estimation model, the proposed method is generic, independent of any particular choice of the baseline.
The structure learning module can be readily added to other prevalent models \cite{newell2016stacked,girdhar2018detect} to improve pose estimation performance, which we intend to explore in future work. 



\newpage

{\small
\bibliographystyle{ieee}
\bibliography{egbib}
}

\end{document}